\documentclass{article}
\usepackage[nonatbib,final]{nips_2017}

\usepackage[utf8]{inputenc} 
\usepackage[T1]{fontenc}    
\usepackage{hyperref}       
\usepackage{url}            
\usepackage{booktabs}       
\usepackage{amsfonts}       
\usepackage{nicefrac}       
\usepackage{microtype}      
\usepackage{enumitem}

\usepackage{graphicx}
\usepackage[font=small]{caption}
\usepackage{subcaption}
\graphicspath{{figures/pdf/}}
\DeclareGraphicsExtensions{.pdf}

\usepackage{tikz}
\usetikzlibrary{bayesnet}

\usepackage{amsfonts}
\usepackage{amsmath}
\usepackage{amsthm}

\usepackage{algorithm}
\usepackage{algorithmic}

\usepackage{booktabs}
\usepackage{multirow}
\usepackage{tabularx}
\usepackage{wrapfig}

\usepackage[style=numeric,
            sorting=none,
            uniquelist=false,
            maxnames=2,
            maxbibnames=10,
            backend=bibtex,
            natbib=true,
            url=true,
            doi=true,
            eprint=true]{biblatex}
\usepackage{macro}
\usepackage{comment}
\usepackage{xcolor}

\DeclareCaptionLabelFormat{andtable}{\tablename~\thetable~\&~#1~#2}

\title{Personalized Survival Prediction with \\
Contextual Explanation Networks}

\author{%
  Maruan~Al-Shedivat \\
  Carnegie Mellon University \\
  \texttt{alshedivat@cs.cmu.edu} \\
  \And
  Avinava~Dubey \\
  Carnegie Mellon University \\
  \texttt{akdubey@cs.cmu.edu} \\
  \And
  Eric~P.~Xing \\
  Carnegie Mellon University \\
  \texttt{epxing@cs.cmu.edu}%
}

\begin{document}

\maketitle

\begin{abstract}
Accurate and transparent prediction of cancer survival times on the level of individual patients can inform and improve patient care and treatment practices.
In this paper, we design a model that concurrently learns to accurately predict patient-specific survival distributions and to explain its predictions in terms of patient attributes such as clinical tests or assessments.
Our model is flexible and based on a recurrent network, can handle various modalities of data including temporal measurements, and yet constructs and uses simple explanations in the form of patient- and time-specific linear regression.
For analysis, we use two publicly available datasets and show that our networks outperform a number of baselines in prediction while providing a way to inspect the reasons behind each prediction.
\end{abstract}

\section{Introduction}

In survival analysis, the goal is to estimate the occurrence time and the risk of an unfavorable event in the future (e.g, death of a patient) that can inform our decisions at present time (e.g., help to select a treatment).
The classical models for this task are the Aalen's additive model~\citep{aalen} and the Cox's proportional hazard model~\citep{cox}, which linearly regress attributes of a patient to the hazard function.
While suitable for comparing populations of patients, these models were not designed for patient-specific prediction.
By reformulating survival analysis as a multi-task classification problem, \citet{yu2011learning} show that a set of temporally ordered linear classifiers provides much more accurate predictions.

Here, we follow the same classification approach and show that using deep learning methods further improves predictive performance on survival data.
While promising, straightforward use of neural networks leads to black-box predictors that lack transparency offered by the linear models.
To overcome this issue, we employ \emph{contextual explanation networks}~\citep[CEN,][]{alshedivat2017cen}---a class of models that learn to predict by generating and leveraging intermediate explanations.
Explanations here are defined as instance-specific simple (linear) models that not only help to interpret predictions but are selected by the network to make predictions for each patient at each time interval.
CENs can be based on arbitrary deep architectures and can process a variety of input data modalities while interpreting predictions in terms of selected attributes.
As we demonstrate in experiments, this approach attains both the best performance as well as interpretability.

\section{Background}

%
First, we present the setup used by \citet{yu2011learning}.
%
The data is represented by patient-specific attributes, $\Xv$, and the times of the occurance of event, $\Tv$.
These times are converted into $m$-dimensional binary vectors, $\Yv := (y^1, \dots, y^m)$, that indicate the corresponding follow up time.
If the death occurred at time $t \in [t_{i}, t_{i+1})$, then $y^j = 0,\, \forall j \leq i$ and $y^k = 1,\, \forall k > i$.
If the data point was censored (i.e., we lack information for times after $t \in [t_{i}, t_{i+1})$), the targets $(y^{i+1}, \dots, y^m)$ are regarded as latent variables.
Note that only $m + 1$ sequences are valid, i.e., assigned non-zero probability by the model, which allows to write the following linear model:
\begin{equation}
    \label{eq:CRF}
    \prob{\Yv = (y^1, \dots, y^m) \mid \xv, \Thetav} = \frac{\exp\left(\sum_{t=1}^m y^t \xv^\top \thetav^t\right)}{\sum_{k=0}^m \exp\left(\sum_{t=k+1}^m \xv^\top \thetav^t\right)}
\end{equation}
The model is trained by optimizing a regularized log likelihood w.r.t. $\Thetav := \{\thetav^t\}_{t=1}^m$.
After training, we get a set of linear models, one for each time interval, used for predicting the survival probability.

\section{Contextual Explanation Networks for Survival Analysis}


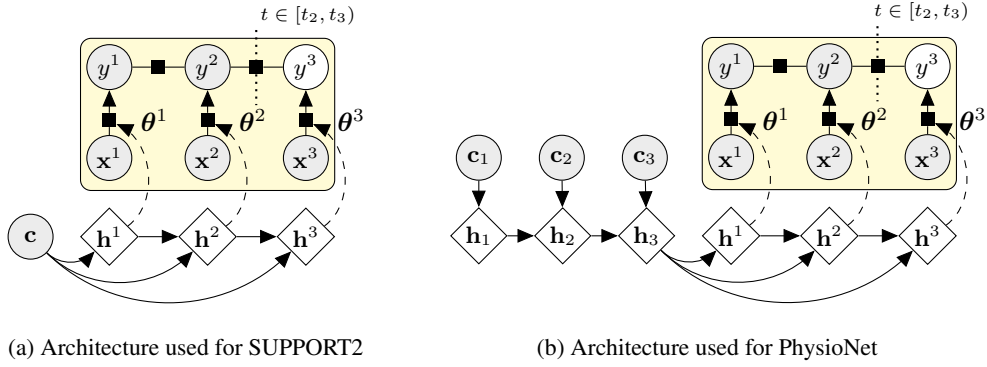
\begin{figure}[t]
    \centering
    \begin{subfigure}[b]{0.49\textwidth}
        \centering
        \begin{tikzpicture}[
            inner/.style={draw, fill=yellow!20, thin, inner sep=2pt},
        ]
            \tikzstyle{latent} = [circle,fill=white,draw=black,inner sep=1pt,
            minimum size=17pt, font=\fontsize{9}{9}\selectfont, node distance=1]

            \node (c) [obs] {$\cv$};
            \node (x_) [obs, transparent, right=10pt of c] {};

            \node (h4) [det, right=10pt of c] {$\hv^1$};
            \node (h5) [det, right=15pt of h4] {$\hv^2$};
            \node (h6) [det, right=15pt of h5] {$\hv^3$};

            \edge[bend right=45] {c} {h4};
            \edge[bend right=45] {c} {h5};
            \edge[bend right=45] {c} {h6};

            \edge {h4} {h5};
            \edge {h5} {h6};

            \node (x1_) [obs, transparent, above=10pt of h4] {};
            \node (x3_) [obs, transparent, above=10pt of h6] {};
            \node (y3_) [obs, transparent, above=17pt of x3_] {};
            \plate[inner, inner sep=2pt] {CEN} {(x1_)(x3_)(y3_)} {};

            \node (x1) [obs, above=10pt of h4] {$\xv^1$};
            \node (x2) [obs, above=10pt of h5] {$\xv^2$};
            \node (x3) [obs, above=10pt of h6] {$\xv^3$};

            \node (y1) [obs, above=17pt of x1] {$y^1$};
            \node (y2) [obs, above=17pt of x2] {$y^2$};
            \node (y3) [latent, above=17pt of x3] {$y^3$};

            \factor[above=3pt of x1]{x1-y1} {} {} {};
            \node (theta1) [const, right=9pt of x1-y1] {$\thetav^1$};
            \factoredge {x1} {x1-y1} {y1};
            \factor[above=3pt of x2]{x2-y2} {} {} {};
            \node (theta1) [const, right=9pt of x2-y2] {$\thetav^2$};
            \factoredge {x2} {x2-y2} {y2};
            \factor[above=3pt of x3]{x3-y3} {} {} {};
            \node (theta1) [const, right=9pt of x3-y3] {$\thetav^3$};
            \factoredge {x3} {x3-y3} {y3};

            \factor[right=7pt of y1]{y1-y2} {} {} {};
            \factoredge[-] {y1} {y1-y2} {y2};
            \factor[right=7pt of y2]{y2-y3} {} {} {};
            \factoredge[-] {y2} {y2-y3} {y3};

            \edge[bend right=60, dashed] {h4} {x1-y1};
            \edge[bend right=60, dashed] {h5} {x2-y2};
            \edge[bend right=60, dashed] {h6} {x3-y3};

            \draw[dotted, thick] (3.0, 2.75) -- (3.0, 1.75);
            \node[anchor=west, font=\fontsize{7}{8}\selectfont, inner sep=0pt] at (3.05, 2.95) {$t \in [t_2, t_3)$};
        \end{tikzpicture}
        \caption{Architecture used for SUPPORT2}\label{fig:MLP-CRF}
    \end{subfigure}
    \begin{subfigure}[b]{0.49\textwidth}
        \centering
        \begin{tikzpicture}[
            inner/.style={draw, fill=yellow!20, thin, inner sep=2pt},
        ]
            \tikzstyle{latent} = [circle,fill=white,draw=black,inner sep=1pt,
            minimum size=17pt, font=\fontsize{9}{9}\selectfont, node distance=1]

            \node (c1) [obs] {$\cv_1$};
            \node (c2) [obs, right=14pt of c1] {$\cv_2$};
            \node (c3) [obs, right=14pt of c2] {$\cv_3$};
            \node (x_) [obs, transparent, right=10pt of c3] {};

            \node (h1) [det, below=10pt of c1] {$\hv_1$};
            \node (h2) [det, right=10pt of h1] {$\hv_2$};
            \node (h3) [det, right=10pt of h2] {$\hv_3$};

            \node (h4) [det, right=10pt of h3] {$\hv^1$};
            \node (h5) [det, right=15pt of h4] {$\hv^2$};
            \node (h6) [det, right=15pt of h5] {$\hv^3$};

            \edge {c1} {h1};
            \edge {c2} {h2};
            \edge {c3} {h3};

            \edge {h1} {h2};
            \edge {h2} {h3};
            \edge[bend right=45] {h3} {h4};
            \edge[bend right=45] {h3} {h5};
            \edge[bend right=45] {h3} {h6};
            \edge {h4} {h5};
            \edge {h5} {h6};

            \node (x1_) [obs, transparent, above=10pt of h4] {};
            \node (x3_) [obs, transparent, above=10pt of h6] {};
            \node (y3_) [obs, transparent, above=17pt of x3_] {};
            \plate[inner, inner sep=2pt] {CEN} {(x1_)(x3_)(y3_)} {};

            \node (x1) [obs, above=10pt of h4] {$\xv^1$};
            \node (x2) [obs, above=10pt of h5] {$\xv^2$};
            \node (x3) [obs, above=10pt of h6] {$\xv^3$};

            \node (y1) [obs, above=17pt of x1] {$y^1$};
            \node (y2) [obs, above=17pt of x2] {$y^2$};
            \node (y3) [latent, above=17pt of x3] {$y^3$};

            \factor[above=3pt of x1]{x1-y1} {} {} {};
            \node (theta1) [const, right=9pt of x1-y1] {$\thetav^1$};
            \factoredge {x1} {x1-y1} {y1};
            \factor[above=3pt of x2]{x2-y2} {} {} {};
            \node (theta1) [const, right=9pt of x2-y2] {$\thetav^2$};
            \factoredge {x2} {x2-y2} {y2};
            \factor[above=3pt of x3]{x3-y3} {} {} {};
            \node (theta1) [const, right=9pt of x3-y3] {$\thetav^3$};
            \factoredge {x3} {x3-y3} {y3};

            \factor[right=7pt of y1]{y1-y2} {} {} {};
            \factoredge[-] {y1} {y1-y2} {y2};
            \factor[right=7pt of y2]{y2-y3} {} {} {};
            \factoredge[-] {y2} {y2-y3} {y3};

            \edge[bend right=60, dashed] {h4} {x1-y1};
            \edge[bend right=60, dashed] {h5} {x2-y2};
            \edge[bend right=60, dashed] {h6} {x3-y3};

            \draw[dotted, thick] (5.30, 1.75) -- (5.30, 0.75);
            \node[anchor=west, font=\fontsize{7}{8}\selectfont, inner sep=0pt] at (5.25, 1.95) {$t \in [t_2, t_3)$};
        \end{tikzpicture}
        \caption{Architecture used for PhysioNet}\label{fig:LSTM-CRF}
    \end{subfigure}%
    \caption{%
    {\CEN} architectures used in our survival analysis experiments.
    Context encoders were time-distributed single hidden layer MLP (a) and LSTM (b) that produced inputs for another LSTM over the output time intervals (denoted with $\hv^1$, $\hv^2$, $\hv^3$ hidden states respectively).
    Each hidden state of the output LSTM was used to generate the corresponding $\thetav^t$ that were further used to construct the log-likelihood for CRF.
    }\label{fig:CEN-CRF}
\end{figure}

Here, we take the same structured prediction approach but consider a slightly different setup.
In particular, we assume that each data instance (patient record) is represented by three variables: the \emph{context}, $\Cv$, the \emph{attributes}, $\Xv$, and the \emph{targets}, $\Yv$.
Our goal is to learn a model, $\prob[\wv]{\Yv \mid \Xv, \Cv}$, parametrized by $\wv$ that can predict $\Yv$ from $\Xv$ and $\Cv$.
Note that inputs have two representations, $\Xv$ and $\Cv$, where $\Xv$ is a set of attributes that will be used to interpret predictions\footnote{%
It is common to have the data to be of multiple representations some of which are low-level or unstructured (e.g., image pixels, sensory inputs), and other are high-level or human-interpretable (e.g., categorical variables).
To ensure interpretability, we would like to use deep networks to process the low-level representation (the \emph{context}) and construct explanations as \emph{context-specific probabilistic models} on the high-level features.}.
Contextual explanation networks (CENs) are defined as models that assume the following form:
\begin{equation}
    \label{eq:CEN-general}
    \Yv \sim \prob{\Yv \mid \Xv, \thetav}, \quad
    \thetav \sim \prob[\wv]{\thetav \mid \Cv}, \quad
    \prob[\wv]{\Yv \mid \Xv, \Cv} = \int \prob{\Yv \mid \Xv, \thetav} \prob[\wv]{\thetav \mid \Cv} d\thetav
\end{equation}
where $\prob{\Yv \mid \Xv, \thetav}$ is a predictor parametrized by $\thetav$.
Such predictors are called \emph{explanations}, since they explicitly relate interpretable variables, $\Xv$, to the targets, $\Yv$.
The conditional distribution $\prob[\wv]{\thetav \mid \Cv}$ is called the \emph{context encoder} processes the context representation, $\Cv$, and generates parameters for the explanation, $\thetav$.

For survival analysis, we want explanations to be in the form of linear CRFs as given in \eqref{eq:CRF}.
Hence, our contextual networks with CRF-based explanations are defined as follows:
\begin{equation}\label{eq:CEN-CRF}
    \begin{aligned}
        & \thetav^t \sim \prob[\wv]{\thetav^t \mid \Cv},\, t \in \{1,\dots, m\}, \quad
        \Yv \sim \prob{\Yv \mid \Xv, \thetav^{1:m}}, \\
        & \prob{\Yv = (y^1, y^2, \dots, y^m) \mid \xv, \thetav^{1:m}} \propto
        \exp\left\{\sum_{t=1}^m y^i (\xv^\top \thetav^t) +  \omega(y^t, y^{t + 1})\right\} \\
        & \prob[\wv]{\thetav^t \mid \Cv} := \delta(\thetav^t, \phi^t_{\wv, \Dv}(\cv)), \quad \phi^t_{\wv, \Dv}(\cv) := \alphav(\hv^{t})^\top \Dv, \quad \hv^t := \mathrm{RNN}(\hv^{t-1}, \cv)
    \end{aligned}
\end{equation}
A few things to note here.
First, the model generates explanations for each patient and for each time interval.
Second, depending on the nature of the context representation, $\Cv$, {\CENs} process it and generate $\thetav^t$ for each time step using a recurrent encoder (Figure~\ref{fig:CEN-CRF}).
We use a deterministic RNN-based encoder, $\phi^t$, that selects parameters for explanations from a global dictionary, $\Dv$, using soft attention (for details on dictionary-based context encoding, see~\citep{alshedivat2017cen}).
Finally, the potentials between attributes, $\xv$, and targets, $y^{1:m}$, are linear functions parameterized by $\thetav^{1:m}$; the pairwise potentials between targets, $\omega(y_i, y_{i + 1})$, ensure that configurations $(y_i = 1, y_{i+1} = 0)$ are improbable (i.e., $\omega(1, 0) = -\infty$ and $\omega(0, 0) = \omega_{00}$, $\omega(0, 1) = \omega_{01}$, $\omega(1, 1) = \omega_{10}$ are learnable parameters).
Given these constraints, the likelihood of an uncensored event at time $t \in [t_j, t_{j+1})$ is
\begin{equation}
    \label{eq:CRF-uncensored}
 \prob{T = t \mid \xv, \Thetav} = \exp\left\{\sum_{i=j}^m \xv^\top \thetav^i\right\} \Bigg{/} \sum_{k=0}^m \exp\left\{\sum_{i=k+1}^m \xv^\top \thetav^i\right\}
\end{equation}
and the likelihood of an event censored at time $t \in [t_j, t_{j+1})$ is
\begin{equation}
    \label{eq:CRF-censored}
    \prob{T \geq t \mid \xv, \Thetav} = \sum_{k=j+1}^m \exp\left\{\sum_{i=k+1}^m \xv^\top \thetav^i\right\} \Bigg{/} \sum_{k=0}^m \exp\left\{\sum_{i=k+1}^m \xv^\top \thetav^i\right\}
\end{equation}
The joint log-likelihood of the data consists of two parts:
\begin{equation}
    \Lc(\Yv, \Xv; \Thetav) = \sum_{i \in \text{NC}} \prob{T = t_i \mid \xv_i, \Thetav} + \sum_{j \in \text{C}} \prob{T > t_j \mid \xv_j, \Thetav}
\end{equation}
where $\text{NC}$ is the set of non-censored instances (for which we know the outcome times, $t_i$) and $\text{C}$ is the set of censored instances (for which only know the censorship times, $t_j$). The objective is optimized using stochastic gradient descent. See \citep{alshedivat2017cen} for more details.



\section{Experiments}\label{sec:survival-analysis}

In our experiments, we consider the datasets, models, and metrics as described below.
We compare CENs with a number of baselines quantitatively as well as visualize the learned explanations.

\textbf{Datasets.}
We use two publicly available datasets for survival analysis of of the intense care unit (ICU) patients:
(a) SUPPORT2\footnote{\url{http://biostat.mc.vanderbilt.edu/wiki/Main/DataSets}.}, and
(b) data from the PhysioNet 2012 challenge\footnote{\url{https://physionet.org/challenge/2012/}.}.
The data was preprocessed and used as follows:
\begin{itemize}[itemsep=2pt,topsep=0pt,parsep=1pt,leftmargin=2em]
    \item \texttt{SUPPORT2:}
    The data had 9105 patient records and 73 variables.
    We selected 50 variables for both $\Cv$ and $\Xv$ features.
    Categorical features (such as \texttt{race} or \texttt{sex}) were one-hot encoded.
    The values of all features were non-negative, and we filled the missing values with -1.
    For CRF-based predictors, the survival timeline was capped at 3 years and converted into 156 discrete intervals of 7 days each.
    We used 7105 patient records for training, 1000 for validation, and 1000 for testing.
    \item \texttt{PhysioNet:}
    The data had 4000 patient records, each represented by a 48-hour irregularly sampled 37-dimensional time-series of different measurements taken during the patient's stay at the ICU.
    We resampled and mean-aggregated the time-series at 30 min frequency.
    This resulted in a large number of missing values that we filled with 0.
    The resampled time-series were used as the context, $\Cv$, while for the attributes, $\Xv$, we took the values of the last available measurement for each variable in the series.
    For CRF-based predictors, the survival timeline was capped at 60 days and converted into 60 discrete intervals.
\end{itemize}


\begin{table}[t!]
    \centering
    \caption{\small%
    Performance of the classical Cox and Aalen models, CRF-based models, and {\CENs} that use LSTM or MLP for context embedding and CRF for explanations.
    The numbers are averages from 5-fold cross-validation; the std. are on the order of the least significant digit.
    @K denotes the temporal quantile, i.e., the time point such that K\% of the patients in the data have died or were censored before that point.}
    \fontsize{8}{10}\selectfont
    \def\arraystretch{1.2}
    \begin{tabular}[t]{@{}lrrrr|lrrrr@{}}
        \toprule
        \multicolumn{5}{c|}{\textbf{SUPPORT2}} &
        \multicolumn{5}{c}{\textbf{PhysioNet Challenge 2012}} \\
        \midrule
        \textbf{Model} &
        \textbf{Acc@25} & \textbf{Acc@50} & \textbf{Acc@75} & \textbf{RAE} &
        \textbf{Model} &
        \textbf{Acc@25} & \textbf{Acc@50} & \textbf{Acc@75} & \textbf{RAE} \\
        \midrule
        Cox         & $84.1$    & $73.7$    & $47.6$    & $0.90$    &
        Cox         & $93.0$    & $69.6$    & $49.1$    & $0.24$    \\
        Aalen       & $87.1$    & $66.2$    & $45.8$    & $0.98$    &
        Aalen       & $93.3$    & $78.7$    & $57.1$    & $0.31$    \\
        CRF         & $84.4$    & $89.3$    & $79.2$    & $0.59$    &
        CRF         & $93.2$    & $85.1$    & $65.6$    & $0.14$    \\
        MLP-CRF     & $\mathbf{87.7}$    & $89.6$    & $80.1$    & $0.62$    &
        LSTM-CRF    & $93.9$    & $86.3$    & $68.1$    & $\mathbf{0.11}$    \\
        \midrule
        MLP-CEN     & $85.5$    & $\mathbf{90.8}$    & $\mathbf{81.9}$    & $\mathbf{0.56}$    &
        LSTM-CEN    & $\mathbf{94.8}$    & $\mathbf{87.5}$    & $\mathbf{70.1}$    & $\mathbf{0.09}$    \\
    %
        \bottomrule
    \end{tabular}
    \label{tab:performance-survival}
    \vspace{-1.5ex}
\end{table}

\textbf{Models.}
For baselines, we use the classical Aalen and Cox models and the CRF from \citep{yu2011learning}, where all used $\Xv$ as inputs.
Next, we combine CRFs with neural encoders in two ways:
\begin{itemize}[itemsep=2pt,topsep=0pt,parsep=1pt,leftmargin=2em]
    \item[(i)]
    We apply CRFs to the outputs from the neural encoders (denoted MLP-CRF and LSTM-CRF, all trainable end-to-end).
    Similar models have been show very successful in the natural language applications~\citep{collobert2011natural}.
    Note that parameters of the CRF layer assign weights to the latent features and are no longer interpretable in terms of the attributes of interest.
    \item[(ii)] We use {\CENs} with CRF-based explanations, that process the context variables, $\Cv$, using the same neural networks as in (i) and output parameters for CRFs that act on the attributes, $\Xv$.
\end{itemize}


\begin{figure}[t]
\centering
\includegraphics[width=\textwidth]{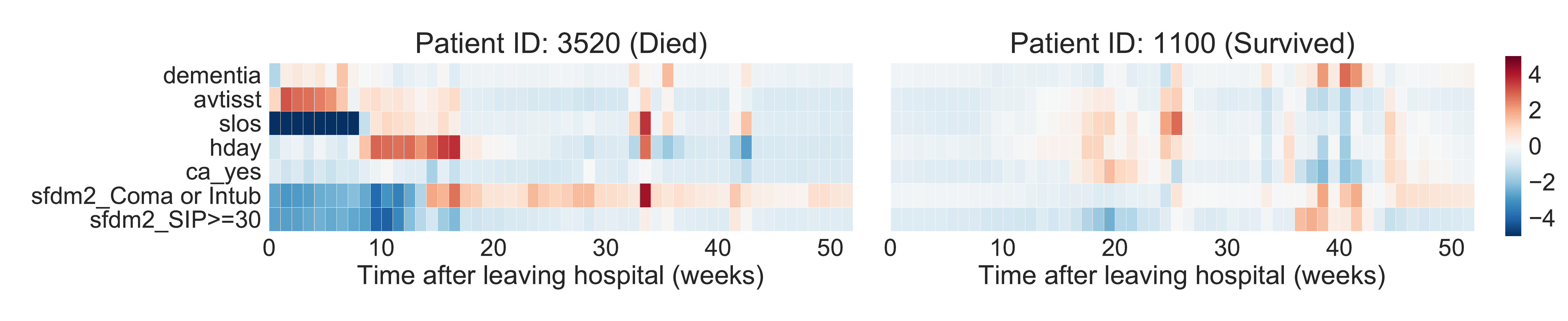}%
\caption{%
Weights of the CEN-generated CRF explanations for two patients from SUPPORT2 dataset for a set of the most influential features: \texttt{dementia} (comorbidity), \texttt{avtisst} (avg. TISS, days 3-25), \texttt{slos} (days from study entry to discharge), \texttt{hday} (day in hospital at study admit), \texttt{ca yes} (the patient had cancer), \texttt{sfdm2 Coma or Intub} (intubated or in coma at month 2), \texttt{sfdm2 SIP} (sickness impact profile score at month 2).
Higher weight values correspond to higher feature contributions to the risk of death after a given time point.
}
\label{fig:support2-heatmaps}
\end{figure}


\begin{wrapfigure}[17]{l}{0.36\textwidth}
\vspace{1ex}
\centering
\includegraphics[width=0.36\textwidth]{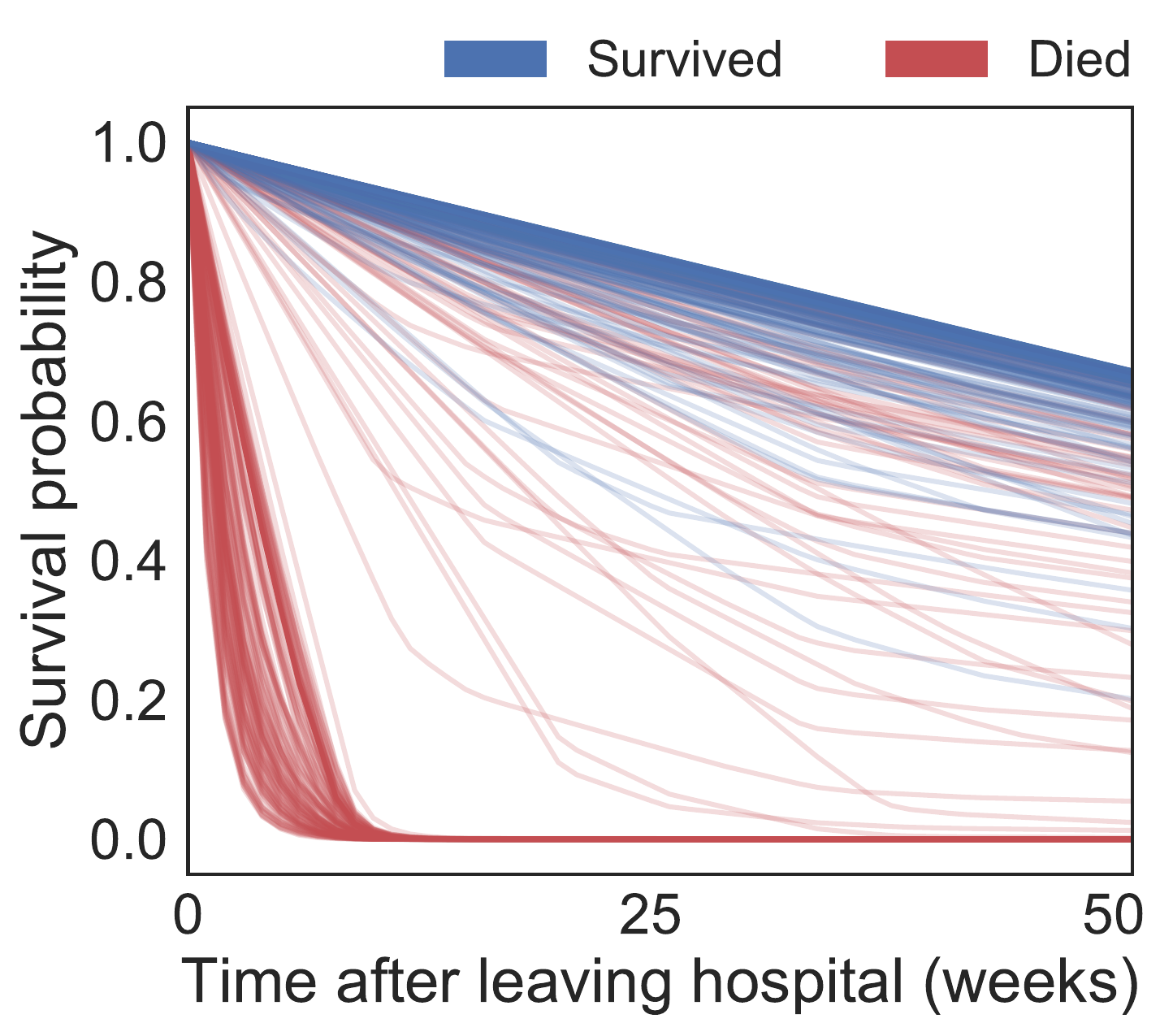}%
\caption{%
CEN-predicted survival curves for 500 random patients from SUPPORT2 test set.
Color indicates death within 1 year after leaving the hospital.
}\label{fig:support2-lifelines}
\end{wrapfigure}

\textbf{Metrics.}
Following \citet{yu2011learning}, we use two metrics specific to survival analysis:
(a) accuracy of correctly predicting survival of a patient at times that correspond to 25\%, 50\%, and 75\% population-level temporal quantiles (i.e., time points such that the corresponding percentage of the patients in the data had their time of the last follow up prior to that due to censorship or death) and
(b) the relative absolute error (RAE) between the predicted and actual time of death for non-censored patients.

\textbf{Quantitative results.}
The results for all models are given in Table~\ref{tab:performance-survival}.
Our implementation of the CRF baseline reproduces (and even slightly improves) the performance reported by~\citet{yu2011learning}.
CRFs built on representations learned by deep networks (MLP-CRF and LSTM-CRF models) improve upon the plain CRFs but, as we noted, can no longer be interpreted in terms of the original variables.
On the other hand, {\CENs} outperform neural CRF models on certain metrics (and closely match on the others) while providing explanations for the survival probability predictions for each patient at each point in time.

\textbf{Qualitative results.}
To inspect predictions of {\CENs} qualitatively, for any given patient, we can visualize the weights assigned by the corresponding explanation to the respective attributes at each time interval.
Figure~\ref{fig:support2-heatmaps} shows explanation weights for a subset of the most influential features for two patients from SUPPORT2 dataset who were predicted as survivor and non-survivor.
These explanations allow us to better understand patient-specific temporal dynamics of the contributing factors to the survival rates predicted by the model (Figure~\ref{fig:support2-lifelines}).
This information can be used for model diagnostics (i.e., help us understand whether we can trust a particular prediction) and as more fine-grained information useful for decision support.

\setlength\bibitemsep{2.5\itemsep}
\printbibliography



\end{document}